\documentclass{article}
\usepackage{acra}
\usepackage{graphicx}
\usepackage{svg}
\usepackage{amsmath}
\usepackage{amssymb}
\usepackage{appendix}
\usepackage{tabularx}
\usepackage{url}
\usepackage[colorinlistoftodos,prependcaption,textsize=tiny]{todonotes}

\title{\LARGE \bf
Exploring GPT-4 for Robotic Agent Strategy with Real-Time State Feedback and a Reactive Behaviour Framework}

\author{Thomas O'Brien \and Ysobel Sims \\ University of Newcastle, Australia \\ 
{\tt\small \{thomas.obrien,ysobel.sims\}@uon.edu.au}}
\begin{document}

\maketitle
\thispagestyle{empty}
\pagestyle{empty}


\begin{abstract}

We explore the use of GPT-4 on a humanoid robot in simulation and the real world as proof of concept of a novel large language model (LLM) driven behaviour method. LLMs have shown the ability to perform various tasks, including robotic agent behaviour. The problem involves prompting the LLM with a goal, and the LLM outputs the sub-tasks to complete to achieve that goal. Previous works focus on the executability and correctness of the LLM's generated tasks. We propose a method that successfully addresses practical concerns around safety, transitions between tasks, time horizons of tasks and state feedback. In our experiments we have found that our approach produces output for feasible requests that can be executed every time, with smooth transitions. User requests are achieved most of the time across a range of goal time horizons. 

\end{abstract}

\section{Introduction}

The need to interact with, transform and navigate the physical world to realise our objectives is considered a significant factor in the cognitive evolution of the human brain \cite{Wolpert2011}. Therefore, the quest to design robotic agents capable of manipulating objects and navigating their environment at a human-equivalent level remains one of the most ambitious research pursuits in AI and robotics \cite{Kuipers2017}. This monumental challenge is evident when considering the diverse and intricate ways humans execute tasks. Recent work has demonstrated that robots can emulate human actions in daily tasks like cooking a meal or loading a dishwasher \cite{Tedrake2019}. While these feats are undeniably remarkable, they often rely on highly controlled conditions and deploying the robot robustly in an unknown setting remains an unattainable goal. The humanoid form factor offers a potential advantage in that many human-centric tools, equipment and environments are prevalent in the world, thus, a humanoid robot is highly practical and adaptable for a range of different tasks in these settings \cite{ReherAmes2021}.

Large language models (LLMs) have inspired researchers across the globe. Transformer-based architectures trained on a large corpus of data have exhibited emergent behaviours~\cite{wei2022emergent}, where they reason and perform actions beyond their training data. Recent work~\cite{SayCan2022,InnerMonologue,ProgPrompt,codeaspolicies2022} has investigated the feasibility of using such models in an agent's strategy and planning layers. This problem involves the model taking a general goal such as `clean the soda spill' and outputting tasks for that goal such as `grab the soda can', `place the soda can in the recycling bin', `grab the sponge' and `clean the spill with the sponge'.

An agent may be programmed to reason about how to achieve high-level goals using logic or machine learning techniques such as reinforcement learning. However, it may be infeasible for these methods to work for any scenario an agent may encounter. Service robots in homes, care facilities or commercial environments will encounter unique scenarios they must navigate. Developers may not anticipate these scenarios. A robotic agent that shuts down and cannot assist when the problem is outside of its original scope is less valuable than a general-purpose robot that can devise a solution for any scenario.

Large language models such as LLaMA~\cite{touvron2023llama} and GPT-4~\cite{openai2023gpt4} have shown natural language skills and appear to understand various topics. Existing works have demonstrated the possibility of using such models on robotic agents in the real world~\cite{yu2023language,ProgPrompt}, where the agent can adapt and plan for unseen environments. With the release of increasingly more impressive language models and research into their use in agent planning, we get closer to a future where LLMs become embodied and can have direct physical impact on their environment.

This work explores a practical setup involving a real-time message-passing software system combined with a complete behaviour framework driven with an LLM. The LLM determines the current tasks in a tree-based behaviour system for the robot to perform to achieve the user's request. We consider long-horizon goals with high-frequency state feedback from our real-time message-passing system, which re-prompts the model online for a rolling set of actions over time. 

Our contributions are

\begin{itemize}
    \item Use of an innovative tree-based behaviour system to facilitate large language models in real robotic scenarios, prioritising seamless transitions and safety
    \item Successful exploration of frequent feedback state polling to update the model's immediate task requests
    \item Implementation of large language model behaviour on a humanoid robot
    \item Experimentation of the performance of large language models on tasks with different time horizons
\end{itemize}

\section{Background}

We consider the literature relevant to robotic agent strategy and provide an overview of large language models.

\subsection{Related Work}

\textbf{Task planning in robotics} aims to generate a sequence of high-level actions to achieve a goal within an environment. There exist numerous conventional methods in the literature that leverage domain-specific language representations, such as Answer Set Programming (ASP)~\cite{Gelfond2014} and Planning Domain Definition Language (PDDL)~\cite{Ghallab1998} combined with complex heuristics-based search techniques~\cite{Blai2001} to obtain solutions. The necessity to define tasks and the surrounding environment through these languages, coupled with the requirement of complex heuristics, constrains the adaptability of these methods when applied to more expansive settings or increasingly complicated tasks. Learning-based methods, including hierarchical and deep-reinforcement learning~\cite{Ceola2019} approaches, have shown promise; however, they have large data requirements and can fail to scale effectively~\cite{MAIMON2000175}.

\textbf{Task planning with LLMs} involves translating user requests in natural language into high-level sequences of actions and has become an emergent trend in robotics. Most literature provides the model with the general scene information before starting, such as the types of objects in the scene~\cite{Huang2022}. Some works provide environment feedback after the execution of each sub-task~\cite{SayCan2022}. More complex systems consider whether tasks failed~\cite{InnerMonologue}. ProgPrompt~\cite{ProgPrompt} uses asserts during execution to close the loop; however, it fails to do this during real robot experiments due to uncertainty in the robot's world model. In SayPlan~\cite{rana2023sayplan}, a 3D scene graph is utilised for planning and representing scene information; however, the framework is limited by its requirement for a pre-constructed 3D scene graph and its assumption that objects will remain unchanging after the creation of the map. This limitation substantially narrows its ability to adapt to changing real-world conditions. Feasibility of the LLM output has become a concern in this area. Some works~\cite{SayCan2022,xie2023translating} have issues with the ability of the output to run; however, ProgPrompt~\cite{ProgPrompt} solves this using programmatic prompts. Research into the horizon time of user requests is lacking. There is a lack of goals with extremely long time horizons, such as `keep the kitchen clean' or `play soccer'. ProgPrompt's `make dinner'~\cite {ProgPrompt} is an example of a longer request with multiple possible solutions.

\textbf{Reactive and modular approaches in robotics} aim to create systems that are highly responsive to environmental stimuli, prioritising immediate action over complex, long-term planning. Unlike traditional methods that rely on creating a detailed plan before execution, these approaches allow robots to adapt in real-time, making them well-suited for dynamic and uncertain settings. Finite State Machines, the subsumption architecture~\cite{Brooks1986}, and behaviour trees~\cite{Colledanchise2018} are among this category's most widely used paradigms. In these approaches, the robot's actions are guided directly by sensory input, and complex behaviours emerge from the interaction of more straightforward, individual behaviours and skills. They are often used in applications where quick, real-time responses to changing conditions are more important than executing a pre-defined plan. However, the practical implementation of this architecture has proven to be complex, particularly for general-purpose robots where the decoupling of high-level and low-level behaviours can be challenging.

\textbf{Reactive and modular approaches combined with LLMs} is a new and relatively unexplored area, with potential advantages over task planners and vanilla reactive-based approaches. In the work of \cite{lykov2023llmbrain} and \cite{cao2023robot}, a LLM is used to generate behaviour trees from text descriptions, which yields promising results; however, they are limited to the case of pre-computing the tree offline. To the best of the authors' knowledge, no work polls the model during execution at a high frequency to obtain a rolling task plan, which adjusts a reactive methods structure based on dynamic changes in the environment and a high level user request. In this work, we look to expand upon this category by leveraging LLMs to dynamically construct a reactive task layer in real-time. 

\subsection{Large Language Models}
In the field of natural language processing, text generation is a problem that can be described as a language model using a conditional probability distribution $P(y\mid x; \theta)$, where $x$ is the input text, $y$ is the output text generated and $\theta$ are model parameters. Large language models are a form of text generation using transformer-based architectures trained over huge-scale corpus, often from the internet. An example is the OpenAI GPT-3 model, which has 175 billion parameters~\cite{brown2020language}. These state-of-the-art large language models have shown strong capability in adaptation and generalisation, which has shifted the traditional ``pre-train-then-fine-tune'' paradigm to the new ``prompt-based-learning'' paradigm~\cite{reynolds2021prompt}. 

In the context of robotic-agent planning, most techniques attempt to directly use existing models, such as OpenAI's GPT-3 and GPT-4, without further training~\cite{Huang2022,InnerMonologue,xie2023translating,ProgPrompt}, while others do consider fine-tuning~\cite{SayCan2022,brohan2023rt2} to improve performance. Training data, particularly for real robots, is hard to obtain, so while the latter offers a reliable way to get useful results, the former provides flexibility and generalisation, where the method can potentially be plugged into any agent-based system.

\section{Methods}

In this section, we provide an overview of NUClear~\cite{Houliston2016}, a system framework, and the Director~\cite{director2023}, a behaviour framework. We outline how an LLM is integrated into these frameworks and leveraged to dynamically construct a reactive task layer in real-time to achieve a higher level user request. Additionally, we outline how the behaviour framework used in combination with an LLM promotes correctness and safety.

\subsection{Software Architecture}

We use NUClear~\cite{Houliston2016}, a modular, low-latency, real-time message-passing architecture for resource-constrained robotics systems. NUClear shares several similarities with the widely-used Robot Operating System (ROS) \cite{Macenski_2022}, where both offer a modular middleware framework for robotics built upon the idea of a publisher-subscriber model. Modular architectures break down complex systems into manageable, independent components with the benefits of readability, writability and easy reusability.

\subsection{Behaviour Framework}

\begin{figure}[t]
    \centering
    \includegraphics[width=0.35\textwidth]{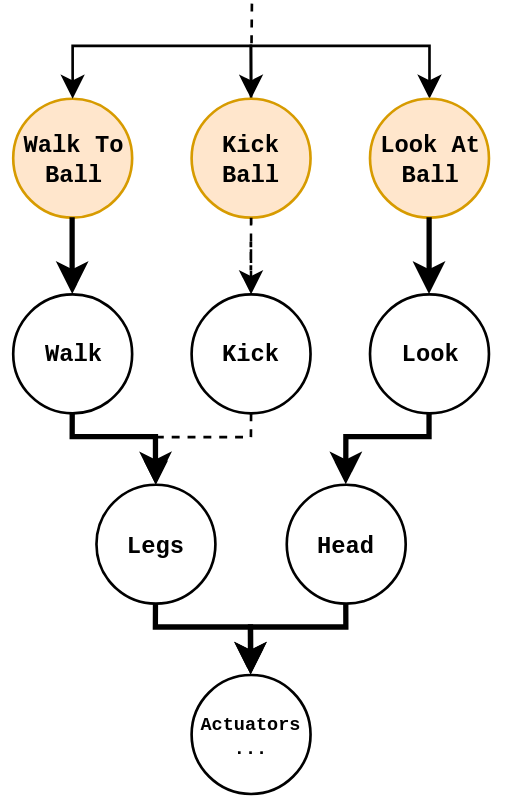}
    \caption{Example of a Director tree for walking to and kicking a soccer ball~\protect\cite{director2023}}
    \label{fig:Director}
\end{figure}

The Director is an algorithm and framework for controlling a system's behaviour flow. It is implemented as a NUClear extension and exploits the modularity of the publisher-subscriber model. It uses Providers (publishers/subscribers) and Tasks (messages) to build a tree-like structure, where Tasks request functionality, and Providers provide the functionality for those behaviours. It allows low-level behaviours and skills to be developed in an independent fashion, combining benefits of both behaviour trees and subsumption architectures. See \cite{director2023} for more details. While the Director effectively enables straightforward development of behaviours and functionalities like this, the system's significant limitation is its inability to generalise to more abstract, high-level and context-dependant tasks. Its architecture becomes very cumbersome when attempting to tackle the intricacies involved in general-purpose task solving and execution, which is the primary motivation for combining the Director framework with an LLM. 

Figure~\ref{fig:Director} shows an example of a Director tree for a humanoid robot tasked with walking to and kicking a ball. At the top layer, highlighted in solid orange, are higher level tasks such as ``Walk  To Ball''. These tasks are well suited for an LLM to dynamically select in order to achieve a higher level request. This idea is the primary contribution of this work and will be discussed further in the following section.

\subsection{LLM Integration}

We use the previously described aspects of the software and behaviour framework in combination with a large language model, tasked with selecting the required Tasks to achieve a high level user request. The language model exists within the system as a module (Provider) that runs at a fixed frequency.  It is prompted with potential tasks the agent can perform, expected output format, state information, and a goal. The exact prompt used for experiments is provided in Appendix~\ref{appendix}. To facilitate straightforward programmatic interpretation, the output of the model is formatted in a specific structure. Previous work has shown that programmatic structure results in output that is executable~\cite{ProgPrompt}, as the model restricts itself to only using the available functions. The specific output form of the model is

\begin{verbatim}
    Task: <task> Priority: <priority>
\end{verbatim}

\begin{figure}[h!]
    \centering
    \includegraphics[width=0.5\textwidth]{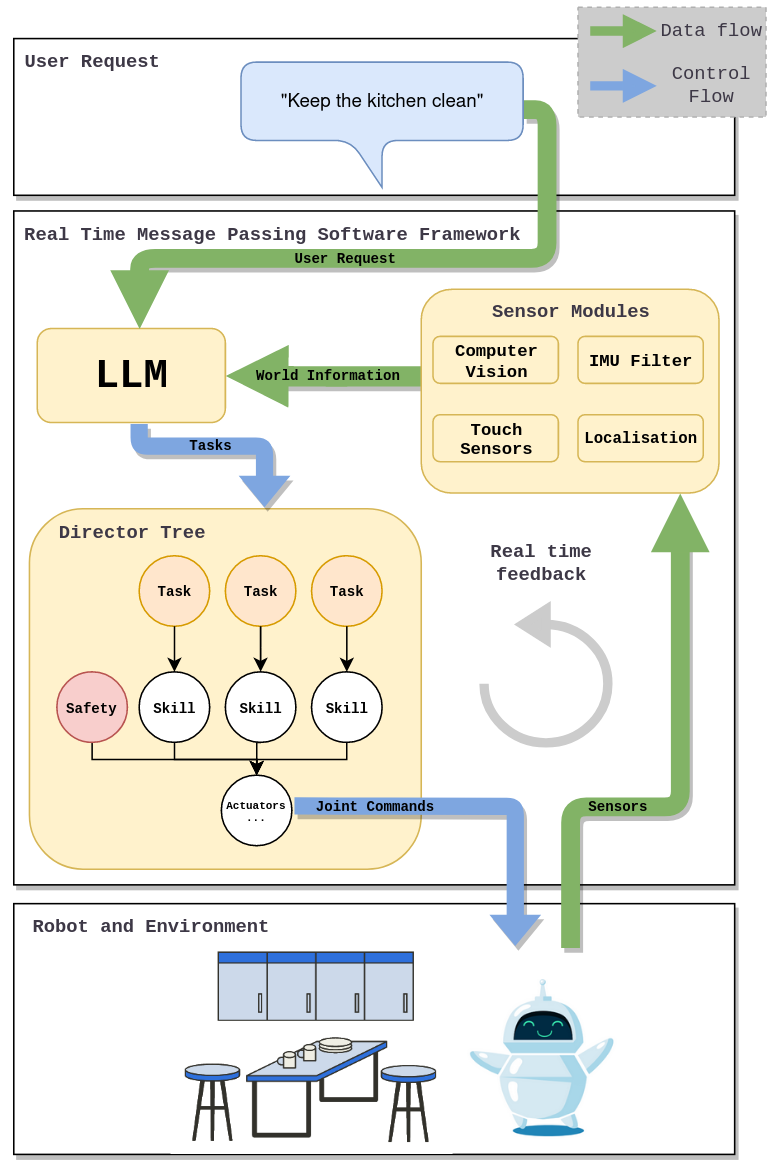}
    \caption{Overview of method with flow of data and control between user request, the system and the robot. The user request and state information is used to create the LLM prompt. The LLM outputs tasks to the Director system, which creates joint commands at the actuation level to move the robot. Safety module/s may take over the actuation level to ensure safety compliance.}
    \label{fig:1}
\end{figure}

The text is processed with a regular expression to extract each task and priority. For example, if the model outputs ``Task: Wave Priority: 2'', the LLM Provider should request the sub-task Wave with a priority of 2. Our method requires the existing implementation of lower-level modules (Providers) that execute functionality for the tasks emitted. A diagram illustrating the entire system is shown in Figure~\ref{fig:1}.

The Director framework has the following beneficial properties which allows the LLM to be safely implemented on a robot.

\textbf{Resource acquisition} is a feature of the Director where only one Provider can access a particular resource at a time, for example a motor, and the Provider with access is determined based on priority. Resource acquisition prevents different subsystems from fighting over the same resource, ensuring smooth execution. This ensures that the output of the LLM does not result in a conflict of resources that prevents the robot from working. 

\textbf{Soft transitions} is the other major advantage of the Director framework. When transitioning from one motion to another requires the robot to move into a safe state before it transitions (critical in bipedal locomotion), the Director enables this through conditions put on Providers, which stop those Providers from running unless a condition is met. For example, Provider A executes a particular motion but will not run unless the safe stability state condition is met. Provider B can cause this safe stability condition to occur. The Director will prioritise Provider B to run so that the condition is met and Provider A can run. Therefore, the output of the LLM does not need to be concerned with transitions between tasks, as the Director system handles this naturally.

Other Providers can run alongside the language model Provider as siblings. These Providers can provide safety guarantees for the robot and act with a higher priority. One example used in our work is a falling and getup management strategy to account for the humanoid robot falling over. This subsystem will take over from the language model when necessary. Other possible safety-orientated subsystems include a balance Provider to prevent falls and a Provider that avoids walking into people and objects. In high-risk real applications, safety guarantees are crucial for the uptake of machine learning methods.

\section{Experiments}
\begin{figure}[!h]
    \centering
    \includegraphics[width=0.5\textwidth]{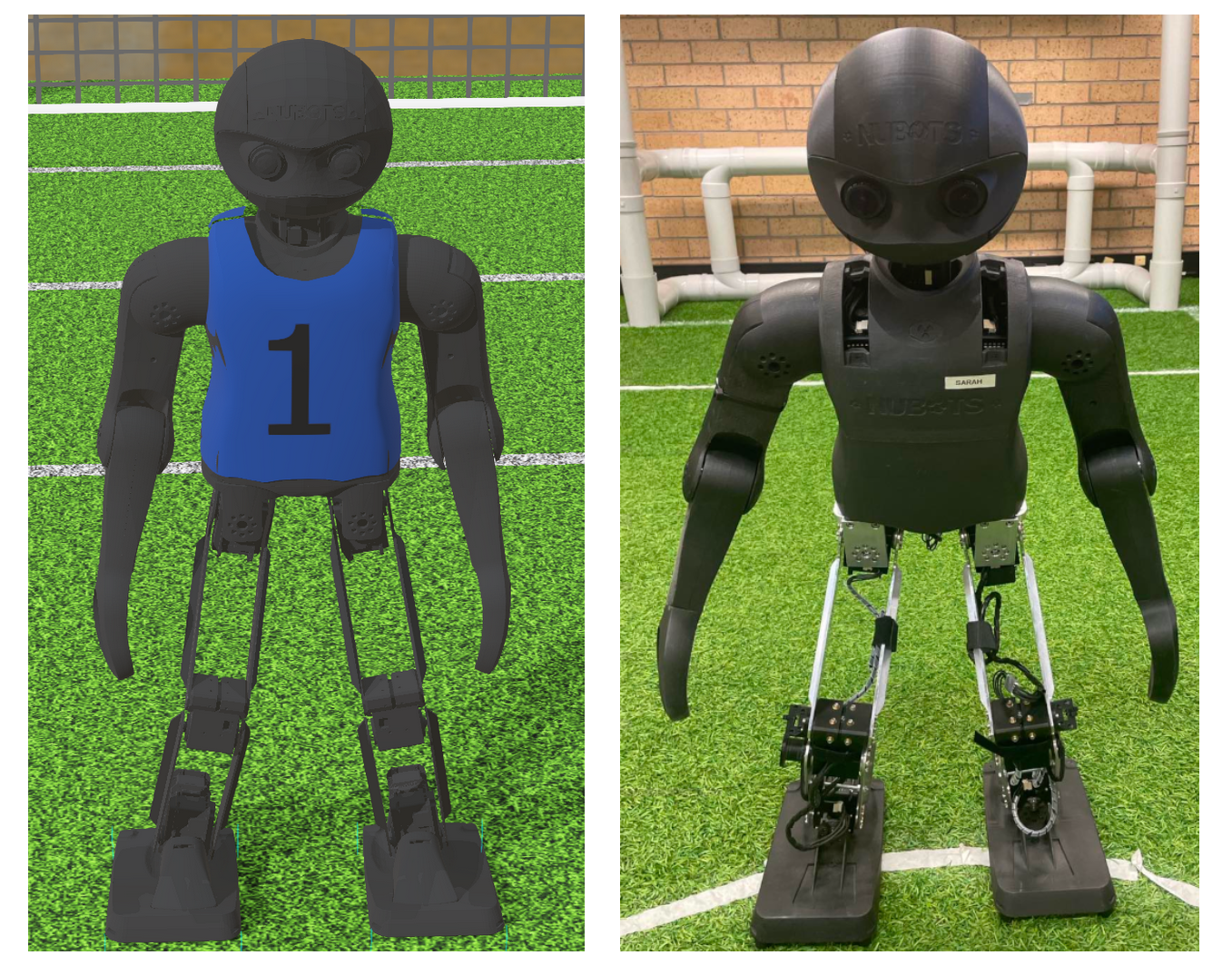}
    \caption{NUgus platform in simulation environment Webots (left) and real hardware (right).}\label{fig:nugus}
\end{figure}

We conduct pilot experiments for our proposed method in simulation and on a real humanoid robot platform designed to play soccer. The simulation environment used is Webots, and we use the NUgus humanoid platform based on the igus Humanoid Open Platform~\cite{allgeuer2016igus}. The NUgus is a humanoid robot made with aluminium and 3D printed Onyx and with an Intel NUC12WSHi7, OpenCR board, twenty Dynamixel MX-series servos and two FLIR Blackfly S cameras. It has Wi-Fi support, which allows communication with the language model through the OpenAI API. Polling of the IMU and servos runs at 90Hz. Figure~\ref{fig:nugus} shows both the virtual and physical robot.

In the following experiments, a Mahony filter is used for pose estimation to detect a fallen state. A getup subsystem runs parallel to the LLM to provide a safety guarantee that the robot will stop executing tasks from the LLM and will perform a getup action when it falls. 

The ball detector directly gives state information to the LLM. This subsystem uses the Visual Mesh~\cite{Houliston2018VisualMR} to detect soccer balls, which are then post-processed using geometry and an Unscented Kalman filter to estimate the position of the soccer ball on the field.

The agent has access to the following tasks:

\begin{itemize}
    \item \texttt{WalkToBall}: walks to the ball if visible
    \item \texttt{KickBall}: kicks when close to the ball
    \item \texttt{LookAround}: looks around by moving the head
    \item \texttt{LookAtBall}: looks directly at the ball if visible
    \item \texttt{StandStill}: stands still
    \item \texttt{TurnOnSpot}: turns on the spot
    \item \texttt{Wave}: waves its arm 
\end{itemize}

These call subtasks in the Director system to achieve the desired functionality.

The LLM used is GPT-4 with a zero temperature value. This will result in the model always choosing the highest probability word to output, reducing creativity and randomness. The LLM is prompted for tasks every two seconds over the internet using the OpenAI API. Communication with the OpenAI API occurs within the code run onboard the robot, with full access to the Director framework, the robot's task functionality and state information.

Our experiments consider whether GPT-4 can reach a given goal, whether the model's output is executable, and whether the robot smoothly transitions as tasks change. Problems processing the output indicate that the model did not output in the correct format or with non-existent tasks and we class this as non-executable output. A unique aspect of our work is that we also consider the horizon of goals and run tests over a series of goals that grow in horizon length. A short horizon task might be a motion command such as `jump'. A long horizon task may be `keep the kitchen clean' or `play soccer'. These goals have not been investigated much in the literature and are harder to measure. In this work, we provide an initial investigation into long-horizon goals, and we explore the method's performance across different horizon lengths. The different goals used in our experiments are given in Table~\ref{tab:goal_id}. Identification numbers are given in this table and will be used in the text.

Goals 1, 2, 5 and 6 represent increasingly long horizon length goals. Goal 3 is a continuation of goal 2, where the robot is manually pushing the robot to the ground in order to trigger a transition to getting up from the ground and then back to using the LLM output. As our method is a rolling series of prompts with state feedback, we also consider the case where the goal changes during the execution of the program. In goal 4, the `approach the ball' goal is requested, and after twenty seconds, the request changes to `stand still and wave'. We include goal 7 as an alternative to goal 6 to demonstrate how a slight change to the prompt can significantly enhance the performance. Goals 8 and 9 are infeasible but are included to investigate how the model reacts and whether the output is executable.

\begin{figure}[!h]
    \centering
    \includegraphics[width=0.5\textwidth]{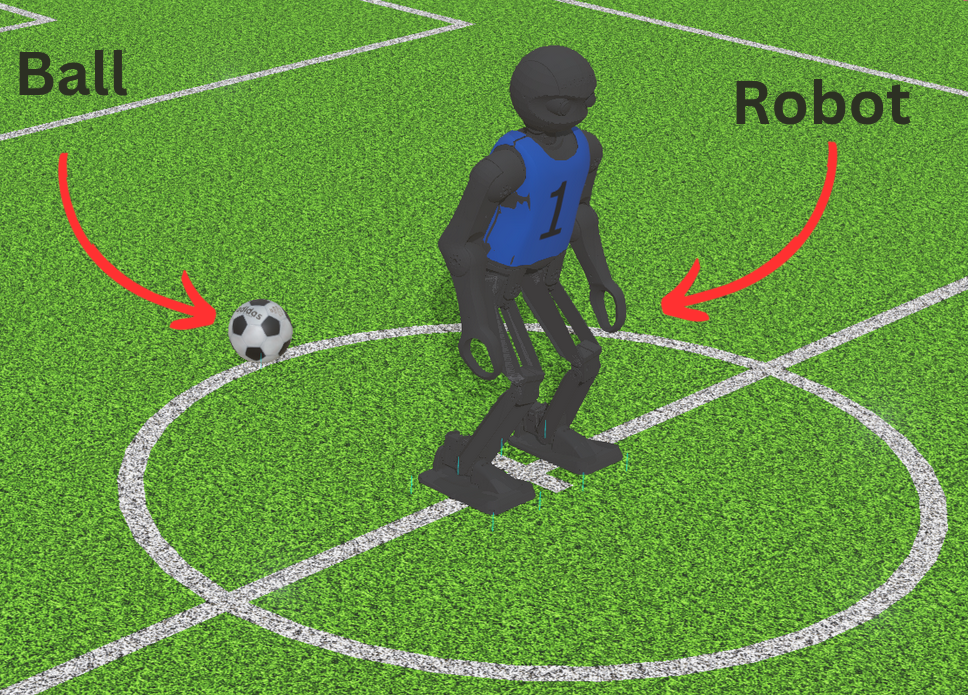}
    \caption{Experiment setup in Webots simulator. Robot is placed facing away from the ball to encourage the use of world information to achieve requests.}\label{fig:experiment}
\end{figure}

For both the simulation and physical experiments, the robot is positioned on the centre line, facing its goalpost. The soccer ball is placed on the centre circle line behind the robot, illustrated in Figure~\ref{fig:experiment}. This positioning requires the robot to explore its environment, as the ball is outside its field of view, and the robot does not know where it is. The program on the robot stops when it either succeeds in its goal or if three runs of the LLM have occurred without the robot progressing towards the given goal. Goals 6 and 7 are satisfied when the robot scores a goal.

\begin{table}[t]\small
    \centering
    \begin{tabular}{c|c}
    \centering
    \textbf{ID} & \textbf{Goal} \\ \hline
    1 & ``Find the ball'' \\ \hline
    2 & ``Approach the ball'' \\ \hline
    3 & ``Approach the ball'' \\&Robot is pushed over during execution  \\ \hline
    4 & ``Approach the ball'' \\& After 20 seconds, ``stand still and wave'' \\ \hline
    5 & ``Approach and kick the ball'' \\ \hline
    6 & ``Play soccer'' \\ \hline
    7 & ``Playing soccer''  \\ \hline
    8 & ``Pick up the ball''  \\ \hline
    9 & ``Jump'' \\ \hline
\end{tabular}
    \caption{Description of each experiment. The ID will be used throughout the text and other tables to reference the different goals.}
    \label{tab:goal_id}
\end{table}

Each experiment is repeated ten times in the simulation and three times on the physical platform.

\section{Results}

We run pilot experiments in simulation in Webots and on a real robot platform. The results demonstrate the potential of our method, with the use of rolling state feedback, a real-time system architecture and a behaviour framework providing a more practical approach for utilising large language models for robotic behaviour.

\subsection{Simulation}

\begin{table}[t]\small
\centering        
    \begin{tabularx}{0.4\textwidth}{>{\centering\arraybackslash}X|>{\centering\arraybackslash}X>{\centering\arraybackslash}X}
        \centering
        \textbf{Goal ID} & \textbf{Success} & \textbf{Executability} \\ \hline
        1 & 0.9 & 1.0 \\ \hline
        2 & 1.0 & 1.0 \\ \hline
        3 & 1.0 & 1.0 \\ \hline
        4 & 1.0 & 1.0 \\ \hline
        5 & 1.0 & 1.0 \\ \hline
        6 & 0.2 & 1.0 \\ \hline
        7 & 0.9 & 1.0 \\ \hline
        8 & - & 1.0 \\ \hline
        9 & - & 0.53 \\ \hline
    \end{tabularx}
    \caption{Results from running the large language model with the given goals in the Webots simulation. Each is repeated ten times. The success column is the rate at which the robot achieves the given goal across runs. Goal 6 and 7 are satisfied when the robot scores a goal. Executability is the rate at which the model's output is feasibly executable on the robot. Success for goal 8 and 9 are not recorded as these requests are not feasible for the robot to complete and are instead used to evaluate the executability of the method in extreme cases.}
    \label{tab:sim_results}
\end{table}

Table~\ref{tab:sim_results} shows the results of the simulation-based experiments. Overall, the requests were satisfied a majority of times. The cases where the LLM could not successfully execute the task were often because the LLM failed to make the robot turn around to find the ball. It would only use the `look around' task in these cases. Without information in the prompt about the robot model and its constraints, the LLM may not realise that `look around' only moves the head within the bounds of human movement and that turning around is necessary. 

The inclusion of goal 8 investigates how the model would react to an infeasible goal. As expected, it does not reach the goal, but it still outputs tasks that were executable by the robot. In this case, the model often chose tasks such as standing still, turning around and looking around. Goal 9 is included for a similar reason but represents a shorter, more direct goal that may push the model to output tasks that are not available. Approximately half of the time, the model's output was not executable due to outputting the task `jump', which is not one of the available tasks. This appeared to be dependent on whether the robot could see the ball or not, since it would attempt to find the ball before trying to jump. 

One run of experiment 7 succeeded in turning around but failed to kick the ball into the goal. However, it would be unfair to judge the LLM on this as it does not have access to localisation information. Without localisation information, the robot does not know if it is kicking towards a goal.

Goal 6, `play soccer', performed poorly as the robot would frequently not turn around. However, goal 7, `playing soccer', proved to be a more reliable prompt. Other works~\cite{Kovalev2022} have also noted a significant change in performance with minor changes to the prompt. We found that simple changes to the prompt, such as added grammatical full-stops, could dramatically change the output. In initial tests with the GPT-3.5-turbo model, `walk to the ball' consistently failed to turn around while goal 2, `approach the ball', succeeded in turning around. GPT-4 did not have this same issue, but a similar situation arose in the play/playing prompts. 

The Director smoothly handled transitions between skills as the LLM changed tasks. The robot did not fall over during standard experiments; therefore, to investigate the safety subsystem's performance, a further experiment involving pushing the robot over is included. The robot enters a fallen state after being manually rotated in the simulation while running goal 2. The Director system switched to the getup subsystem when it detected the fallen state. The robot recovered by standing back up, and the LLM subsystem could take over again to complete the goal. Developing and including the relevant subsystem as a sibling to the LLM can guarantee safety in various situations.

Goal 4 involved switching after twenty seconds from the `approach the ball' goal to the `stand still and wave' goal. The agent performed these requests without issues, cleanly ending the walk to stop and wave. It not only shows the capability of the Director for smooth transitions but also shows that our method allows for changing the prompt online. Due to the system's modular nature, the prompt could change at any time based on the state or user feedback, such as microphone input. The robot does not need to complete its first goal to move on to the next goal, providing flexibility.

Overall, the experiments in simulation successfully demonstrated our proposed method. The LLM responded to state feedback and the rolling nature of prompts adequately. The Director supported the transitions and successfully provided a safety guarantee. Limitations of our experimental setup show the potential for future development around expanding the state feedback and low-level systems to support the LLM further.

\subsection{Real World}

Table~\ref{tab:real_results} shows the results of the physical experiments on the NUgus robotic platform. The LLM functionality is high-level and abstracted away from the physical actuation through the Director tree. Therefore, the performance was similar to the simulation. One noticeable difference was the quality of state feedback on the real robot. 
 
\begin{table}[t]\small
    \centering
    \begin{tabularx}{0.4\textwidth}{>{\centering\arraybackslash}X|>{\centering\arraybackslash}X>{\centering\arraybackslash}X}
    \centering
    \textbf{Goal} & \textbf{Success} & \textbf{Executability} \\ \hline
    1 & 1.0 & 1.0 \\ \hline
    2 & 1.0 & 1.0 \\ \hline
    3 & 1.0 & 1.0 \\ \hline
    4 & 1.0 & 1.0 \\ \hline
    5 & $0.\dot{6}$ & 1.0 \\ \hline
    6 & $0.\dot{6}$ & 1.0 \\ \hline
    8 & - & 1.0 \\ \hline
    9 & - & 0.0 \\ \hline
\end{tabularx}
    \caption{Results from running the large language model with the given goals on the physical robot platform. Each runs three times. The success column is the rate at which the robot achieves the given goal across runs. Goal 6 is satisfied when the robot scores a goal. Executability is the rate at which the model's output is feasibly executable on the robot. Success for goals 8 and 9 are not recorded as these requests are not feasible for the robot to complete and are instead used to evaluate the executability of the method in extreme cases.}
    \label{tab:real_results}
\end{table}

For the experiments with the four main prompts - goals 1, 2, 5 and 6 - the real robot performed similarly to the robot in simulation. Issues with the vision feedback prompted us to simplify the task on the real robot for the final two goals. In these cases, the robot starts facing the ball rather than away from the ball. More often than not, it succeeded in kicking and scoring; however, the robot occasionally missed due to poor path planning and vision. While the LLM provided reasonable tasks, the underlying systems struggled to adequately perform when faced with more uncertain feedback than the simulation robot. These issues show that our methods success depends heavily on the quality of the low-level systems. 

The other prompts - falling while approaching the ball, approaching the ball and then waving, picking up the ball, and jumping - performed similarly on the physical and simulation robots. The only exception was goal 9. The model attempted to run the non-existent `jump' task every time. The reason for this is likely the ball detections. When no detections exist, the robot tries to search for it. Once it has seen the ball, it tries to jump. In the simulation experiments, the robot could sometimes see the ball and other times, it was out of view. The physical robot always had ball detections, whether true or false positive.

\subsection{Limitations and Future Work}

While the outcomes of this study are encouraging, we acknowledge that the present implementation does have a number of limitations, which are listed as follows.

\textbf{Computational Cost} of running the model in the cloud or locally. Currently, even powerful desktop computers lack the resources to load the weights of a large language model and run small models at a fast enough rate to be implemented online in the loop. Affordable robots are resource-constrained and don't have the hardware to run LLMs in real time. Using cloud solutions such as OpenAI's API, as we have done in this work, has monetary costs per token, rate limits and restricts the system to being implemented in an environment with internet access. This area would benefit from research into models that can run locally and produce reasonable output.

\textbf{Volatile Output}, where small changes to the prompt can significantly impact the performance, is noticeable in our experiments. Simple grammatical changes like full stops can affect the LLM's output. For example, the difference between `play soccer' and `playing soccer' was dramatic, with the former failing to succeed most of the time while the latter did the opposite. A possible avenue for optimising prompts comes from recent work using LLMs~\cite{yang2023large}, where user input can be optimised by another LLM to reduce this volatility.  Additionally, with our configuration, the same prompt yields non-deterministic results even at a temperature value of zero. This is probably a consequence of the limited token count in our chosen output format, highlighting the importance in further exploration of ideal output formats.

\textbf{Available Tasks} and the low-level implementation for those tasks. While some works aim to produce end-to-end models, our approach considers integrating a language model into an existing robotic system. 
The quality and availability of low-level behaviours limit the ability to achieve a higher-level request. Furthermore, the ability to encode the tasks within natural language limits an agent's range of functions. Future work involving a more comprehensive range of tasks and experiments would benefit the development and evaluation of this method.

\textbf{World Information} is limited by the existing modules in the system. While our method works well in the cases presented in this paper, more complex tasks involving more diverse environmental feedback may become infeasible and form an essential avenue for future work in this area. A straightforward improvement upon this work is adding localisation information to determine if the LLM can utilise two information sources. Prompt engineering may become increasingly more complicated with the expansion of world information. Our prompt does not describe the robot model itself; however, this context could improve its ability.

\section{Conclusion}

In this paper, we have presented a novel approach for integrating an LLM into an existing robotic system to enable high level task execution. Our method addresses practical concerns around safety, transitions between tasks, time horizons of tasks, and environmental feedback. The LLM was integrated into a reactive behaviour framework to dynamically construct reactive task layers in real-time. Our approach has been successfully tested on a humanoid robotic platform, showcasing the potential of combined reactive, modular, and LLM approaches, offering new perspectives over traditional task planners and reactive approaches.  However, we acknowledge that more comprehensive experiments involving a wider range of tasks and environmental feedback are needed to further evaluate and improve the method. We believe that our work opens up new promising avenues for future research in the area of LLMs and reactive behaviour frameworks for robotic agent strategy.

\section*{Acknowledgments}

The authors acknowledge the financial support of the Australian Government through its Research Training Program and of 4Tel Pty Ltd through their sponsorship of the Newcastle Robotics Laboratory. The authors acknowledge the work of past and previous members of the Newcastle Robotics Laboratory since 2002, who contributed to the hardware and software that this work uses. The authors thank Alexandre Mendes and Stephan Chalup for their reviews and supervision of this work.

\appendix
\section{Appendix}\label{appendix}
The experiments in this work use the following prompt:

``Given desired user request: \textless  \texttt{user request}\textgreater\space and current information of the world: Ball \textless  \texttt{"is visible" | "is not visible"}\textgreater, last seen \textless  \texttt{seconds}\textgreater\space seconds ago \textless  \texttt{distance}\textgreater\space m away from you. You have the ability to WalkToBall (requires visible ball), Kick (needs higher priority than walk), LookAround, LookAtBall (requires visible ball), StandStill, TurnOnSpot, Wave. What tasks should you currently do to achieve the request? \textbackslash{n} Provide your response as a list of with format: Task: \textless  task\textgreater\space Priority: \textless  priority\textgreater\textbackslash{n} where \textless  task\textgreater\space is one of the aforementioned tasks, and  \textless  priority\textgreater\space is an integer above 0. Tasks with a larger priority number will take control over a lower priority task if they are moving the same motors. Only use the tasks required right now to progress towards the request. Use the skills to navigate your environment to achieve the task \textless  \texttt{user request}\textgreater\space, considering that if the ball is not visible you will need to move to find it.\textbackslash{n}''

\nocite{*}

\end{document}